\documentclass[conference]{IEEEtran}
\usepackage{cite}
\usepackage{amsmath,amssymb,amsfonts}
\usepackage{algorithm}
\usepackage{algpseudocode}
\usepackage{booktabs}
\usepackage{graphicx}
\usepackage{textcomp}
\usepackage[export]{adjustbox}
\usepackage{subcaption}
\usepackage{listings}
\usepackage{caption}
\usepackage{color}
\usepackage{multirow}
\usepackage{array}
\usepackage{gensymb}
\usepackage{stfloats}
\usepackage[table]{xcolor}
\usepackage[bookmarks=false]{hyperref}
\def\fixme#1{\typeout{FIXED in page \thepage : {#1}}
\bgroup \color{red}{[FIXME: {#1}]} \egroup}
\def\reply#1{\typeout{REPLY in page \thepage : {#1}}
\bgroup \color{blue}{[REPLY: {#1}]} \egroup}
\def\new#1{\typeout{NEW in page \thepage : {#1}}
\bgroup \color{green}{[NEW: {#1}]} \egroup}

\def\BibTeX{{\rm B\kern-.05em{\sc i\kern-.025em b}\kern-.08em
    T\kern-.1667em\lower.7ex\hbox{E}\kern-.125emX}}

\newcommand{\col}[1]{{\cellcolor[HTML]{#1}}}

\begin{document}

\title{DeepPicarMicro: Applying TinyML to Autonomous Cyber Physical Systems} 

\pagestyle{empty}
\author{Michael Bechtel, QiTao Weng, Heechul Yun\\
  University of Kansas, USA. \\
  \{mbechtel, wengqt, heechul.yun\}@ku.edu \\
}

\maketitle

\begin{abstract}

Running deep neural networks (DNNs) on tiny Micro-controller Units (MCUs) is challenging due to their limitations in computing, memory, and storage capacity. Fortunately, recent advances in both MCU hardware and machine learning software frameworks make it possible to run fairly complex neural networks on modern MCUs, resulting in a new field of study widely known as TinyML. However, there have been few studies to show the potential for TinyML applications in cyber physical systems (CPS). 

In this paper, we present DeepPicarMicro, a small self-driving RC car testbed, which runs a convolutional neural network (CNN) on a Raspberry Pi Pico MCU. We apply a state-of-the-art DNN optimization to successfully fit the well-known PilotNet CNN architecture, which was used to drive NVIDIA's real self-driving car, on the MCU. We apply a state-of-art network architecture search (NAS) approach to find further optimized networks that can effectively control the car in real-time in an end-to-end manner. From an extensive systematic experimental evaluation study, we observe an interesting relationship between the accuracy, latency, and control performance of a system. From this, we propose a joint optimization strategy that takes both accuracy and latency of a model in the network architecture search process for AI enabled CPS. 
 
\end{abstract}

\begin{IEEEkeywords}
Real-time, Autonomous Car, Convolutional Neural Network, Microcontroller, Case Study, TinyML
\end{IEEEkeywords}

\section{Introduction}  \label{sec:intro}

Autonomous cyber physical systems (CPS), such as self-driving cars and drones, are a topic with much interest in recent years. The premise is that by employing recent advances in machine learning (ML) algorithms such as deep neural networks (DNNs), CPS can become more intelligent and safer, which benefits society. 

However, executing DNN models is computationally expensive. This limits their applicability to many CPS with significant size, weight, power, cost, and real-time constraints. Therefore, there are increasing research efforts to reduce the computational requirements of employing DNN models. In particular, many researchers and companies are putting significant effort to support DNNs in tiny micro-controller units (MCUs) 
due to their low cost and low power consumption, 
despite their obvious limitations in terms of available computing and memory resources~\cite{lin2020mcunet,lin2021mcunetv2,banbury2020benchmarking,banbury2021micronets}. 

\begin{table}[ht]
  \centering
  \begin{tabular}{|c|c|c|}
    \hline
    Part                    & Raspberry Pi 4 & Raspberry Pi Pico \\
    \hline
    \multirow{2}{*}{CPU}     & BCM2837                   & RP2040 \\
            & 4x Cortex-A72@1.5GHz      & 2x Cortex-M0+@133MHz \\
    \hline
             & 48BK(I)/32KB(D) L1 cache & \\
    Memory   & 1MB L2 (16-way) L2 cache & 264KB SRAM \\
             & 4GB LPDDR4               & \\
    \hline
    Storage  & 8GB+ micro-SD                    & 2MB Flash \\
    \hline
    Power   & 3A            & $<$100mA  \\
    \hline
  \end{tabular}
  \caption{Comparison of hardware resources on a Raspberry Pi 4 microprocessor and a Raspberry Pi Pico MCU.}
  \label{tbl:mp-vs-mcu}
\end{table}

In this paper, we present DeepPicarMicro, a low-cost autonomous car testbed to study the feasibility of AI enabled CPS on tiny MCUs. 
DeepPicarMicro employs an end-to-end deep learning approach, which utilizes a convolutional neural network (CNN) to directly control the physical plant from the camera based sensory input as 
in our prior work DeepPicar\cite{bechtel2018picar}.
The main difference is that DeepPicarMicro utilizes a Raspberry Pi Pico MCU (a dual Cortex-M0+ MCU) as the main computing platform, which is significantly less capable than the Raspberry Pi 3/4 computing platforms used in the DeepPicar. Note that DeepPicar's Raspberry Pi 3/4 platforms were capable enough to run the full unmodified PilotNet model~\cite{bechtel2018picar}, which was used in NVIDIA's real self-driving car~\cite{Bojarski2016}, in real-time. 
Table~\ref{tbl:mp-vs-mcu} shows the differences between a Raspberry Pi 4 and a Raspberry Pi Pico MCU, the latter of which features orders of magnitude smaller computing power and memory/storage availability.

Using DeepPicarMicro, we want to answer the following questions:
(1) Can we run a full-sized PilotNet on a micro-controller?
(2) How can we find optimized neural network architectures for a target micro-controller?
(3) What are the relationships between accuracy, latency, and control performance of an end-to-end DNN model in controlling CPS?

From an extensive systematic experimental evaluation study, we made the following observations.
First, to our surprise, we find that the full sized PilotNet can run on a Pico MCU using a specialized ML framework, namely Tensorflow-lite micro (TFLM)~\cite{david2021tensorflow}, and standard optimization techniques such as 8-bit quantization. However, the unmodified (except quantization) PilotNet model's latency was more than 3 seconds, which is not acceptable for real-time control of CPS. Clearly, there is a strong need to further optimize the network to be able to run on a tiny MCU. 

Second, we apply a state-of-the-art neural architecture search (NAS) approach~\cite{lin2020mcunet} to find smaller variants of the PilotNet by varying the input resolution, depth, width of the full model. In addition, we also employed depthwise-separable convolutions~\cite{sandler2018mobilenetv2} in place of the standard 2D convolutions to further reduce the latency of the models. As a result, we found many PilotNet variants that meet the real-time constraints of the system and achieve high accuracy. Interestingly, however, we observe that less accurate DNN models with lower latency often performed better in practice than more accurate models with higher latency. Even when we compare similarly accurate models, we observe that lower latency models perform better, even if we set the control frequency of the models to be identical (all meeting the same deadline). 
This is because the model's latency affects the reaction time of the CPS system it controls and the quality of the network's output degrades as the network's input becomes stale. This suggests that in a CPS system, not only a network's accuracy but also its latency must be taken into account to predict the model's true performance. Therefore, the standard NAS approach that treats latency as a constraint may not be ideal to find best performing models. 

Third, we evaluate a simple joint optimization strategy, which uses a normalized sum of the DNN model's latency and accuracy to compare a model's predicted performance in a real CPS system. In both simulation and in real-world experiments, we find the joint optimization strategy is effective in predicting a network's real-world performance in controlling the RC car.

In summary, we make the following contributions:
\begin{itemize}
\item We present DeepPicarMicro, a MCU-based autonomous car testbed that employs a CNN-based end-to-end real-time control loop.
\item We present extensive experimental evaluation results showing the possibility of using MCU for AI enabled CPS. 
\item We propose a simple joint optimization strategy that takes both accuracy and latency of a model in the network architecture search process for AI enabled CPS. 
\end{itemize}

The remainder of the paper is organized as follows.
Section~\ref{sec:background} provides a background on MCUs and TinyML.
Section~\ref{sec:overview} gives an overview of the DeepPicarMicro
testbed and our initial evaluations with it. Section~\ref{sec:nas} describes the first NAS approach we use for finding a TinyML model that can run on the DeepPicarMicro.
Section~\ref{sec:evaluation} presents extensive CPS control performance evaluation results on a real-world environment, in addition to our modified NAS approach.
We review related work in Section~\ref{sec:related} and conclude in Section~\ref{sec:conclusion}.
\section{Background} \label{sec:background}

In this section, we provide background on autonomous vehicles, MCUs, and TinyML. 

\subsection{End-to-End Deep Learning for Autonomous Vehicles}

Self-driving cars have been a topic of increasing interest over the past several years. 
A standard approach is to split the task into multiple specialized sub tasks, such as planning and perception~\cite{autoware}. On the other hand, an end-to-end deep learning approach uses a single neural network to produce control outputs directly from the raw sensor input data, which dramatically simplifies the control pipeline~\cite{Levine2016}. 
First introduced in 1989 by Pomerleau~\cite{Pomerleau1989}, many systems have since employed DNN-based control loops to much success~\cite{Bojarski2016,bechtel2018picar,upennf1tenth,donkeycar,deepracer}.

In a DNN based end-to-end control loop based system, training and inference are typically performed separately. 
In general, training a neural network model is computationally
expensive, so it is often done on more powerful PC systems equipped with hardware accelerators (e.g. GPUs). On the other hand, inference operations require relatively less computing power and can thus be run on smaller embedded platforms. However, on such platforms, the model's inference latency becomes an important factor as many systems and applications have real-time constraints. In this paper, we explore the capability of executing a deep neural network on a small microcontroller platform in real-time.

\subsection{Microcontroller Units (MCUs)}
A microcontroller is a small computer that integrates simple CPU core, SRAM and flash memory into a single integrated chip. 
MCUs are inexpensive and consume very little power, often in the range of milliamps (mA). 
As such, they are used in a wide variety of applications, ranging from toys to cars.
Unlike powerful microprocessors, which typically employ complex operating systems and other runtime frameworks to perform sophisticated tasks, MCUs are designed for relatively simple tasks and often do not employ operating systems.  
This allows MCUs to have far more predictable temporal behaviors than microprocessors, as they do not suffer from non-determinism typically seen in standard OSes (e.g. virtual memory, page faults, etc.).
On the other hand, MCUs have very limited computing, memory, and storage capacity, which pose a challenge for complex applications that require large amount of resources, such as machine learning algorithms. 

\subsection{Tiny Machine Learning (TinyML)}
Recently, there are increasing interests and effort to enable ML applications in MCUs, which is collectively known as Tiny Machine Learning or TinyML for short~\cite{banbury2021micronets,lin2020mcunet,lin2021mcunetv2,banbury2020benchmarking}. 
In TinyML, a major goal is to execute machine learning algorithms, such as DNN models, locally on an MCU, instead of relying on communications with larger PC or cloud that requires high energy consumption and suffers long latency. 
Lately, major MCU vendors as well as big tech companies such as Google have developed machine learning frameworks specially tailored for MCUs.  

In this paper, we primarily use the Tensorflow Lite Micro (TFLM) deep learning framework, which is optimized for neural network inference for MCUs~\cite{david2021tensorflow}. TFLM uses a runtime interpreter architecture for portability and supports a wide range of MCUs, including the MCU we used in this study. TFLM supports 8-bit quantified models, which are converted to C char arrays to be directly compiled for the target MCU. To efficiently utilize the limited memory in a MCU, TFLM uses a single pool of statically allocated memory called an "arena" that holds intermediate buffers and computations~\cite{arm_arena_disc}. The size of the arena it can allocate determines the maximum activation size of the neural network model it can support. 
In this paper, we utilize the TFLM framework to execute a CNN model that controls a small-scale RC car autonomously.
\section{DeepPicarMicro}\label{sec:overview}

DeepPicarMicro is a small self-driving RC car that employs an end-to-end deep learning approach utilizing a deep convolutional neural network (CNN) to directly control the motors from the raw camera input data. Such an end-to-end learning approach has been demonstrated in many prior works, including NVIDIA's real self-driving car DAVE-2\cite{Bojarski2016} as well as in our prior work DeepPicar~\cite{bechtel2018picar}. Our main difference compared to all other prior works is that we realized this CNN based end-to-end control system on a tiny MCU. 

\subsection{Hardware Platform and Track}
\begin{figure}[ht]
  \centering
  \includegraphics[width=.45\textwidth]{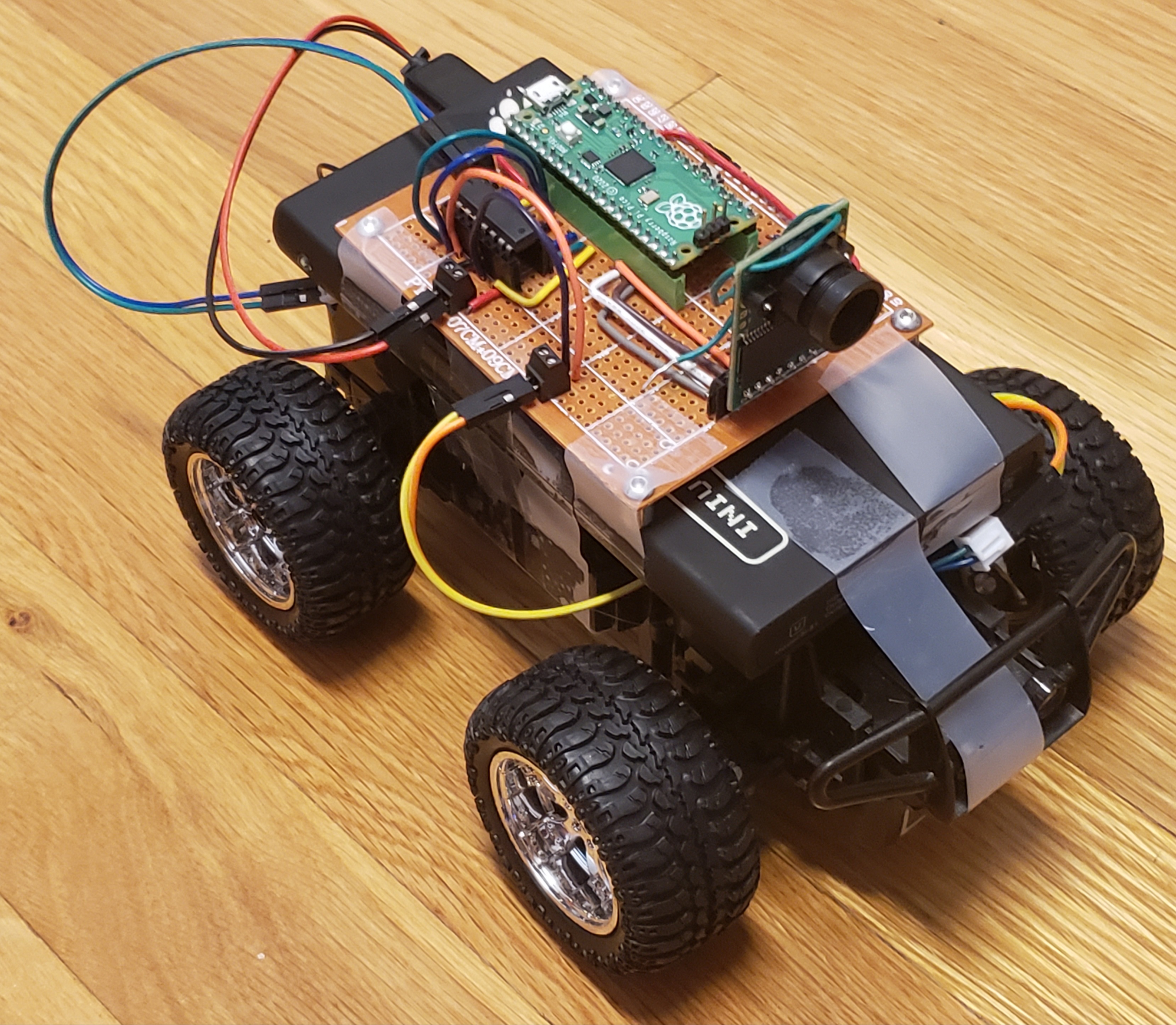}
  \caption{DeepPicarMicro platform.}
  \label{fig:overview}
\end{figure}

\begin{table}[ht]
  \centering
  \begin{tabular}{|c|c|c|}
    \hline
    \textbf{Component}              & DeepPicar         & DeepPicarMicro \\
    \hline
    \textbf{Compute platform}       & Raspberry Pi 3/4              & Raspberry Pi Pico \\ \hline
    \multirow{2}{*}{\textbf{Car platform}}   & New Bright 1:24               & New Bright 1:24  \\
                                    & scale RC car                  & scale RC car \\ \hline
    \textbf{Camera}                 & Playstation Eye               & Arducam Mini 2MP Plus \\ \hline
    \textbf{Motor control}          & Pololu DRV8835            & L293D \\ \hline 
    \textbf{Power source}           & External battery pack         & External battery pack \\ 
    \hline
  \end{tabular}
  \caption{Hardware components used in DeepPicar vs DeepPicarMicro}
  \label{tbl:carbom}
\end{table}

Figure~\ref{fig:overview} shows the DeepPicarMicro, which is comprised of the following components: a Raspberry Pi Pico MCU, a L293D motor driver, an Arducam Mini 2MP Plus, an external battery pack, and a 1:24 scale RC car. Table~\ref{tbl:carbom} shows a comparison of the hardware used in the DeepPicar and DeepPicarMicro platforms.
The camera used on the DeepPicarMicro is able to capture images at a frequency of 7.5Hz ($\sim$133 ms per image),
so we use this as our control frequency. In other words, our control loop has a deadline of 133ms per iteration. Note that the deadline for control systems is often derived from the system's dynamics, though in our case it is determined by hardware limitations. In addition, the camera can only capture images where both the width and height are multiples of four, so we modify all CNN models we find to fit this constraint before loading them onto the DeepPicarMicro.

\begin{figure}[ht]
  \centering
  \includegraphics[width=.45\textwidth]{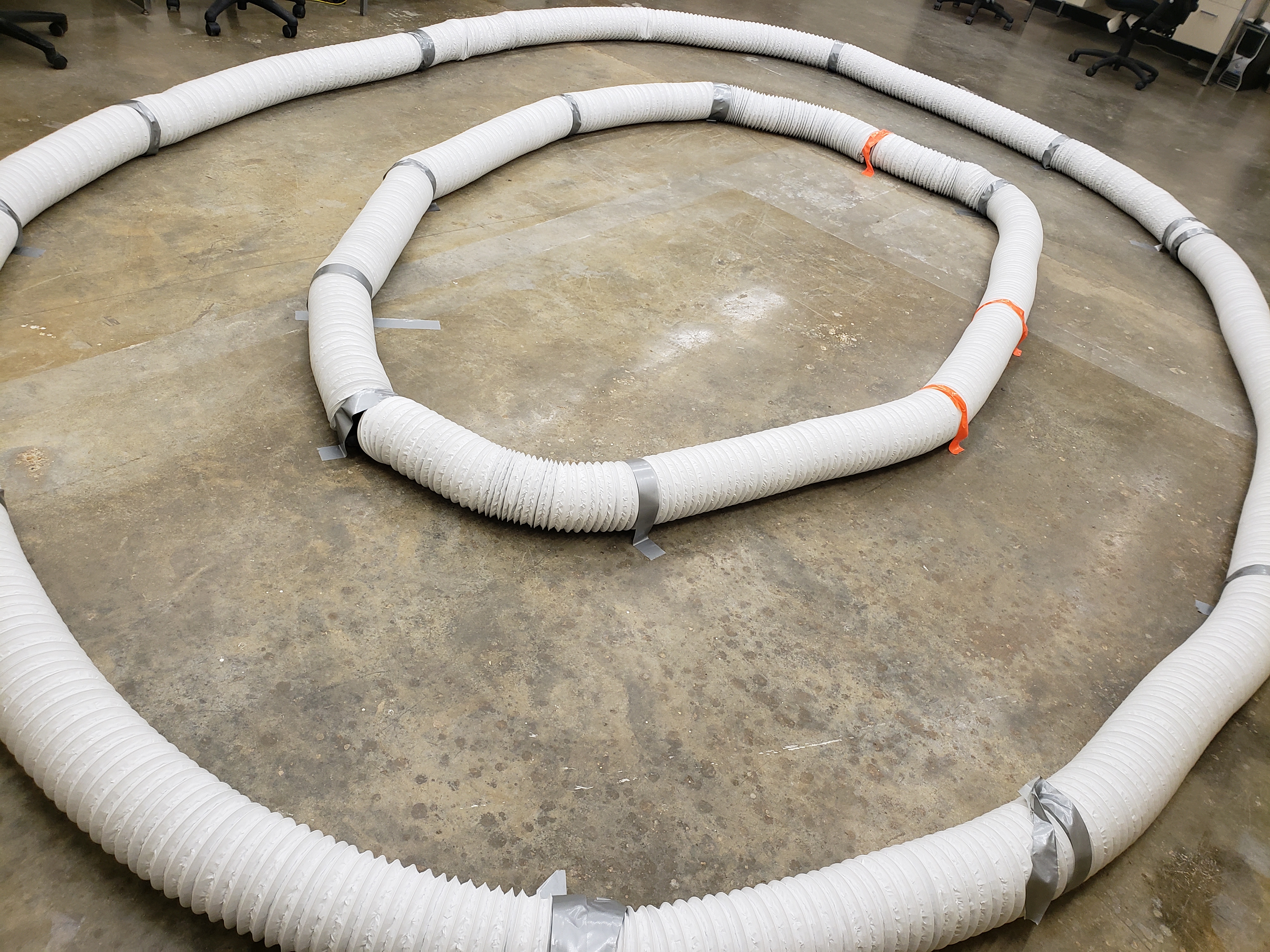}
  \caption{Real-world track used to collect training data and evaluate the control performance of the DeepPicarMicro.}
  \label{fig:track}
\end{figure}

For the environment, we construct a simple track, shown in Figure~\ref{fig:track}, that we use for all of our real-world evaluations in this paper. 

\subsection{The PilotNet Architecture}

For the neural network architecture, we begin by implementing NVIDIA's PilotNet
CNN in TensorFlow using the Keras API. After we train a model, we then use the Tensorflow Lite Micro (TFLM) framework to load and execute the model on the DeepPicarMicro. Additionally, we use integer 8-bit quantization-aware training as the TFLM framework only supports quantized models at the time of writing. 

\begin{table}[ht]
\centering
\begin{tabular}{|c|c|c|c|c|}
    \hline
     Layer  & Input size    & Output size   & Weights   & MACs      \\ \hline
     Conv1  & 66x200x3      & 31x98x24      & 1.8K      & 5.5M      \\ \hline
     Conv2  & 31x98x24      & 14x47x36      & 21.6K     & 14.2M     \\ \hline
     Conv3  & 14x47x36      & 5x22x48       & 43.2K     & 4.8M      \\ \hline
     Conv4  & 5x22x48       & 3x20x64       & 27.7K     & 1.7M      \\ \hline
     Conv5  & 3x20x64       & 1x18x64       & 36.9K     & 663.6K    \\ \hline
     FC1    & 1152          & 100           & 115.3K    & 115.2K   \\ \hline
     FC2    & 100           & 50            & 5.1K      & 5K        \\ \hline
     FC3    & 50            & 10            & 510       & 500       \\ \hline
     FC4    & 10            & 1             & 11        & 51        \\ \hline
     Total  &               &               & 252.2K    & 26.9M     \\ \hline
\end{tabular}
\captionof{table}{PilotNet~\cite{Bojarski2016} architecture}
\label{tbl:pilotnet_layout}
\end{table}

\subsection{Data Collection, Pre-processing and Training}
We manually collect a dataset of 10,000 frame and steering angle pairs around the track, which we will henceforth refer to as the DeepPicarMicro dataset\footnote{We used the DeepPicar platform for data collection as the current iteration of the DeepPicarMicro does not have capability to store the collected data.}. We categorize all steering angles to one of three output classes (left, center, and right) to match the discrete control output space of the DeepPicarMicro. We use 7,500 pairs for training and the remaining 2,500 pairs for validation. To improve the consistency of the training process we also employ the following techniques:
\begin{itemize}
    \item When generating the train and validation sets, we use a constant seed such that the output sets are the same every time they are generated.
    \item We stratify the train and validation sets so that they are both equally proportionate to the output class distribution of the overall dataset.
    \item In our dataset, the majority of the samples are of the car going straight. As such, we perform class balancing so that all three output classes have an equal effect in the changes made to the model's final weight values. Specifically, we assign higher weight values to the left and right output classes and a lower weight to the straight output class.
\end{itemize}
In terms of hardware, the cameras used for the DeepPicar and DeepPicarMicro differ in their image capture properties (e.g. zoom, etc.). As a result, the two cameras will capture different images for the same scene. To account for this during training, we perform an additional translation scheme when pre-processing the input frames.
To be more specific, for each image captured on the DeepPicar, we flip the image and crop it such that the final image closely resembles one captured from the DeepPicarMicro's camera. 

\subsection{Platform Resource Constraints}

To evaluate the performance of the PilotNet model, we initially test three important metrics: memory usage, accuracy, and inference latency. We begin with the memory utilization of the network to determine whether PilotNet with 8-bit quantization can fit on the Pico MCU. We perform a theoretical analysis of PilotNet's per-layer memory usage by calculating the total input and output activation buffer sizes.

\begin{figure}[ht]
\includegraphics[width=0.45\textwidth]{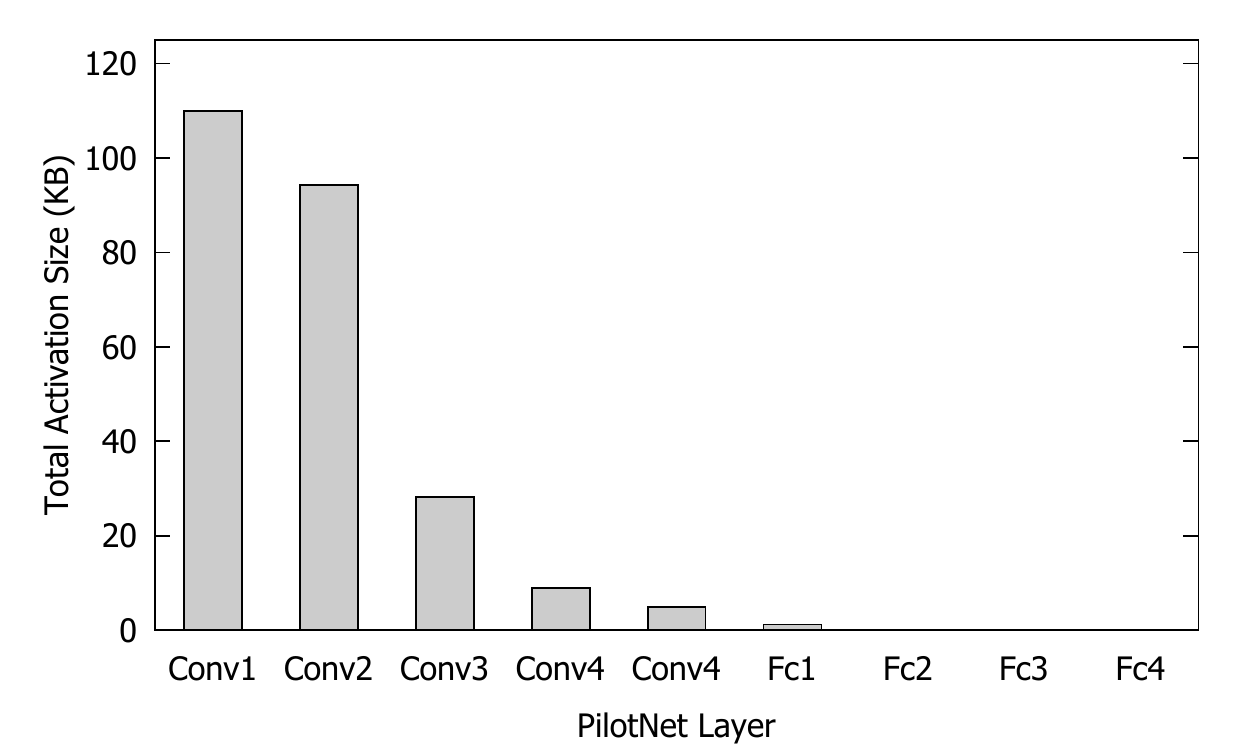}
\caption{SRAM requirement for each PilotNet layer. }
\label{fig:dave2-memory-usage}
\end{figure}

Figure~\ref{fig:dave2-memory-usage} shows the results of this analysis. Importantly, we find that the largest layer only requires ~110KB which easily fits inside the 264KB SRAM available on the Pico MCU. Furthermore, because the TFLM Arena only needs to be as big as the largest layer~\cite{arm_arena_disc}, we find that the full PilotNet model can successfully be run on the Pico. We also find that the initial layers of the network require far more memory than the remainder of the network, which is consistent with prior studies~\cite{lin2021mcunetv2}. 

\subsection{Quantization, Accuracy and Latency}

To test the model's accuracy, we re-feed all 2,500 samples from the validation set to the model. We again categorize the model's predictions to be a left, center, or right output. We then compare the predictions for all images to their respective ground truth outputs and measure the number of samples where the two matched (i.e. the model's prediction was "correct"). Using this method, we test the accuracy for the full PilotNet model, both with and without quantization enabled. Without quantization, PilotNet achieves an accuracy of \emph{87.6\%}, whereas it has an accuracy of \emph{86.9\%} with 8-bit quantization. From this, we find that PilotNet can be quantized and run on the DeepPicarMicro while still achieving comparable accuracy to original 32-bit floating point model.
However, we find that the performance is highly undesirable as it takes over \emph{3 seconds} to process each frame. 

\subsection{Depth-wise Separable Convolutional Layer}
Based on PilotNet's temporal performance, there is little chance it would achieve good control performance. Even though it can run on the Pico with high accuracy, it would still
face significant issues in reacting to external stimuli before the system fails (e.g. the car crashes). To address this, we first try to replace the standard 2D convolutional layers (i.e. Conv2D) with well-known \textit{Depthwise Separable} layers~\cite{chollet2017xception,sandler2018mobilenetv2} to reduce the total multiply-accumulate (MAC) operations. As adopted in many recent TinyML architectures, depthwise separable layers significantly reduce network MACs by separating Conv2D layers into two different operations: (1) a depthwise convolution, and (2) a pointwise convolution. This reduces the computational cost of a convolution from 
\begin{equation}
    C*O_h*O_w*O_d*K^2
\end{equation}
to
\begin{equation}
    C*O_h*O_w*K^2+C*O_h*O_w*O_d
\end{equation}
where $C$ denotes input channels, $O$ denotes output dimensions, and $K$ denotes the kernel size. 
For PilotNet, we replace all five Conv2D layers with equivalent depthwise separable layers and train a new model on the DeepPicarMicro dataset. We henceforth refer to both PilotNet models as either the Conv2D model or Depthwise model, based on the type of convolution they employ. We then perform the same accuracy and inference latency measurement tests for the Depthwise model.

\begin{table}[h!]
\centering
{
    \begin{tabular}{|c|c|c|}
    \hline
                                & \textbf{Conv2D}   & \textbf{Depthwise}     \\ \hline
         \textbf{Weights}       & 252.2K            & 133.7                  \\ \hline
         \textbf{MACs}          & 26.9M             & 2.1M                   \\ \hline
         \textbf{Val. Loss}     & 0.027             & 0.032                  \\ \hline
         \textbf{Accuracy (\%)} & 86.9              & 85.7                   \\ \hline
         \textbf{Latency (ms)}  & 3025              & 525                    \\ \hline
    \end{tabular}
}
\caption{Comparison of PilotNet models with Conv2D and depthwise separable layers.}
\label{tbl:pilotnet_compare}
\end{table}

Table~\ref{tbl:pilotnet_compare} shows the model characteristics for the Conv2D and Depthwise models. 
Notably, the Depthwise model has $\sim$12.7X fewer MACs and $\sim$5.8X faster inference latency compared to the original Conv2D model. 
At the same time, both models have comparable accuracy, with the Depthwise model's accuracy only being $\sim$1\% smaller.

However, we still find the Depthwise model's performance to be unsatisfactory. 
Even with fewer MACs, it still takes the Depthwise model $>$\emph{500 ms} to process a single frame, which is greater than the target 133 ms control period.
Due to this, we next explore the potential for reducing PilotNet's size without overly sacrificing accuracy. For this, we perform a Neural Architecture Search (NAS).

\section{Neural Architecture Search}\label{sec:nas}

In this section, we describe the NAS approach we perform on the PilotNet architecture. 
For the NAS, our goal is to find the model with the highest accuracy while also satisfying the different physical and temporal constraints required by the DeepPicarMicro. In particular, we hold that the model must be small enough to fit in the Pico MCU's SRAM and Flash and that its inference latency be $<$133 ms, the DeepPicarMicro's control period. Importantly, due to the performance gains seen in Table~\ref{tbl:pilotnet_compare}, we perform our NAS on the PilotNet model with depthwise separable layers.

\subsection{Latency Prediction}
While we can calculate the SRAM and Flash usage for a given model layout,
the same can not be said for its inference latency. That is, without profiling a model to measure its latency, we can not directly determine if it meets the 133 ms constraint. However, prior studies have shown that 
a model's MAC operations corresponds to its inference latency~\cite{banbury2021micronets}.  
To find this relationship for the Pico MCU, we run 50 CNN models with MAC operations ranging from $\sim$54.4K to $\sim$2.1M, and measure their respective inference latencies.

\begin{figure}[ht]
  \centering
  \includegraphics[width=.45\textwidth]{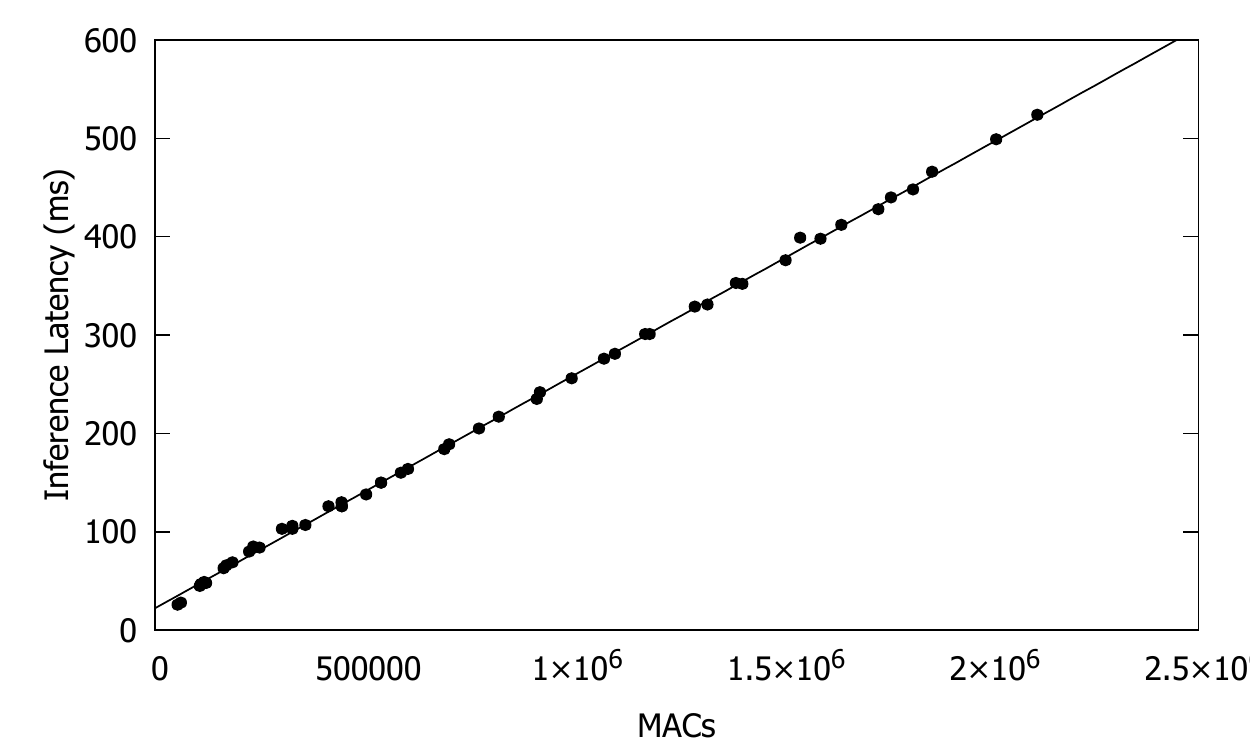}
  \caption{Number of MAC operations vs. inference latency on the Raspberry Pi Pico. We use Linear Regression to find the dotted line of best fit ($R^2$=0.9996).}
  \label{fig:depthwise-regression}
\end{figure}

\begin{equation}
    Latency = 0.000236 * MACs + 22.189388
\label{func:depthwise-regression}.
\end{equation}

Figure ~\ref{fig:depthwise-regression} shows the inference latencies for each model on the Pico MCU. Using these data points, we use Linear Regression to derive Equation~\ref{func:depthwise-regression}, which can be used to predict a model's latency. For example, with this equation an inference latency of 133 ms on the Pico MCU roughly correlates to $\sim$470,000 MACs. Note that this function would not work for models with Conv2D layers because they have have different numbers of operations per convolution overhead~\cite{banbury2021micronets}.

\subsection{Search Space}
Now that we can estimate a model's inference latency, we perform the NAS on PilotNet. We employ a NAS methodology largely influenced by the state-of-the-art MCUNet approach~\cite{lin2020mcunet}. That is, we vary PilotNet's architectural properties that affect the total number of MAC operations. We refer to these properties as \textit{reduction parameters}. However, compared to the MCUNet NAS, we substantially limit the number of network layouts we search to reduce the total execution time of the NAS. Using PilotNet with depthwise separable layers as a backbone, we define a search space with the following two reduction parameters:

\begin{itemize}
    \item \textit{Width} multiplier for all layers, ranging from [0.2, 0.4, 0.7, 0.8, 0.9, 1.0].
    \item Network \textit{Depth}, ranging from a minimum of three layers to maximum of nine layers. We always keep the input convolutional and output fully-connected layers, but vary all unique combinations of the seven middle layers with at least one additional convolutional layer. In total, we evaluate 120 different network layouts in this parameter.
\end{itemize}

In addition, we configure all network layouts to have an input resolution size of 68x68x1. We choose this resolution as it is the smallest that can (1) be used for the full unaltered PilotNet, and (2) be captured by the DeepPicarMicro's camera, as discussed in Section~\ref{sec:overview}. Using this search space, we then perform a two-step NAS. 
First, we construct a model for every network layout in the search space and calculate its number of MAC operations. If the model has $\le$470,000 MACs then we keep that model, otherwise we discard it.
In total, we search 720 different network layouts in the first step, and keep 349 of them. 
In the second step, we train all remaining models on the DeepPicarMicro dataset. For each model, we attempt to train it up to five times to account for random weight initialization. To optimize this process, we define two validation loss thresholds: a target threshold and a fail threshold. At the end of each training iteration, we perform two checks on the model's current validation loss. 
If the validation loss is less than the target threshold, or greater than the fail threshold, then we do not perform any additional training iterations and keep the current model. Likewise, if the model is trained five times without passing either check, then we stop and move on to the next model. 
In our NAS, we use a target threshold of 0.0350, which roughly correlates to 80\% accuracy, and a fail threshold of 0.0450. In the end, we found substantial variance in the validation losses of the 349 models we train, from 0.0287 to 0.0836. In terms of accuracy, this resulted in a range of 62.6\% to 86.6\% accuracy.

\begin{table*}[ht]
\centering
    \begin{tabular}{|c|c|c|c|c|c|c|c|c|c|}
    \hline
        Model \# & Layers & Width & Weights & MACs & Latency (ms) & Val. Loss & Accuracy (\%) & Score & Laps w/o Crash \\ \hline
        1 & 4 & 0.2 & 3.0K & 64.8K & 37 & 0.031 & 84.9 & \col{57bb8a}0.04 & \col{84c78e}7 \\ \hline
        2 & 7 & 0.2 & 26.8K & 151.3K & 58 & 0.031 & 85.0 & \col{87c78e}0.27 & \col{57bb8a}9 \\ \hline
        3 & 8 & 0.7 & 8.8K & 267.6K & 85 & 0.032 & 85.2 & \col{add091}0.50 & \col{84c78e}7 \\ \hline   
        4 & 3 & 0.7 & 5.9K & 214.5K & 73 & 0.042 & 82.1 & \col{b7d392}0.56 & \col{84c78e}7 \\ \hline
        5 & 6 & 0.2 & 1.0K & 70.5K & 39 & 0.060 & 75.3 & \col{bcd493}0.58 & \col{e67c73}0 \\ \hline
        6 & 5 & 0.9 & 13.3K & 358.0K & 107 & 0.033 & 83.8 & \col{dcdc95}0.75 & \col{c8d895}4 \\ \hline
        7 & 7 & 0.8 & 82.7K & 371.4K & 110 & 0.032 & 85.1 & \col{ece097}0.84 & \col{b1d292}5 \\ \hline
        8 & 4 & 0.4 & 1.6K & 128.1K & 52 & 0.073 & 67.7 & \col{fbe498}0.92 & \col{e67c73}0 \\ \hline
        9 & 7 & 0.8 & 9.0K & 315.8K & 97 & 0.050 & 78.2 & \col{ffe398}0.96 & \col{e67c73}0 \\ \hline
        10 & 5 & 0.8 & 5.9K & 312.8K & 96 & 0.052 & 76.5 & \col{fee098}0.98 & \col{f4e399}2 \\ \hline
        11 & 3 & 0.2 & 1.6K & 63.5K & 37 & 0.084 & 65.6 & \col{fedd96}1.00 & \col{e67c73}0 \\ \hline
        12 & 5 & 0.4 & 4.6K & 276.4K & 87 & 0.057 & 75.0 & \col{fdda95}1.03 & \col{f6c28c}1 \\ \hline
        13 & 6 & 0.9 & 53.8K & 421.3K & 122 & 0.044 & 80.4 & \col{f8c78f}1.16 & \col{f4e399}2 \\ \hline
        14 & 6 & 0.7 & 7.0K & 265.5K & 85 & 0.072 & 69.8 & \col{f6be8b}1.23 & \col{e67c73}0 \\ \hline
        15 & 3 & 0.4 & 23.8K & 374.2K & 111 & 0.066 & 72.8 & \col{eb917b}1.56 & \col{e67c73}0 \\ \hline
        16 & 6 & 0.9 & 10.7K & 378.2K & 111 & 0.083 & 62.7 & \col{e67c73}1.71 & \col{e67c73}0 \\ \hline
    \end{tabular}
    \caption{Model statistics and performance for each real-world test case, in order of increasing heuristic score. All models have the same input resolution of 68x68x1, and the latency values are calculated using Equation~\ref{func:depthwise-regression}. }
    \label{tbl:track_perf}
\end{table*}

\subsection{Performance Prediction}

In the process of searching for an optimal TinyML model, most state-of-the-art NAS approaches will optimize their search on a performance based metric, such as accuracy. However, in the context of CPS, it has been shown that inference latency can also have a notable impact on control performance~\cite{we2017functionally,park2021ros2}. Based on these findings, we propose a joint optimization strategy to better predict the control performance of the CNN models we found. In our strategy, we assign a heuristic score to each model that is calculated as follows:

\begin{equation}
    Score = norm(Val Loss) + norm(Latency)
    \label{func:heuristic}
\end{equation}

In this function, we normalize the validation loss and inference latency values for all of the models to be between 0 and 1.
For each model, we then sum the two normalized values together to get a heuristic score between 0 and 2. With this strategy, the intuition is that models with relatively smaller validation losses and inference latencies often perform better. Therefore, the smaller a model's heuristic score is, the better it should perform. Note that we use estimated latencies based on Equation~\ref{func:depthwise-regression} when generating the heuristic scores.

Now that we have CNN models that can effectively run on the DeepPicarMicro in real-time, we next test their control performance in a real-world environment, as well as the effectiveness of our joint optimization strategy.

\section{Evaluation}\label{sec:evaluation}

In this section, we evaluate the control performance of the CNN models from NAS both in a real track and in a simulated track.  

\subsection{Performance in Real Track}

To begin, we perform a more in-depth evaluation of the models found in Section~\ref{sec:nas}. As it would be time consuming and inefficient to test all 349 models, we select a sample subset of 16 models with differing validation losses and inference latencies.
Using Equation~\ref{func:heuristic}, we calculate the heuristic scores for each model. We next evaluate the real-world control performance of each model. For this, we use the DeepPicarMicro testbed on our handmade track from Figure~\ref{fig:track}. 
For each CNN model, we attempt to run it on the DeepPicarMicro for ten individual laps and measure the total number of laps the car is able to finish without crashing. 

Table~\ref{tbl:track_perf} shows the statistics, heuristic scores, and laps completed for each tested model. 
As expected that some CNN models performed very well (green colored) and completed the majority of laps without crashing while some models perform poorly (red colored). An important observation is that a model's accuracy alone was not a sufficient indicator to predict the system's true performance in the track. 
For example, the best model (\#2) we tested completed 9 laps without a single crash, but another similarly accurate---in terms of validation loss and accuracy---model (\#7) was only able to complete 5 labs without crash. Note also that models \#6 and \#7 achieve good accuracy yet perform worse than significantly less accurate model \#4. When we consider latency into account, however, it is clear that these highly accurate models did not work well as their latencies are significantly higher than others. As such, we find that our heuristic score that considers both accuracy and latency into account generally performed well in predicting each model's true relative performance. 
That being said, our joint optimization strategy is not perfect and can incorrectly predict the performance of some models. For example, the model \#5 in the table, which has a relatively good score of 0.58, performed very poorly in the real world and couldn't complete any laps around the track. Unlike other models with low (good) scores, all of which had accuracy of $>$80\%, model \#5 had a relatively low accuracy of $\sim$75\%. This indicates that accuracy is too low, even with a fast inference time, it can lead to undesirable results.

\subsection{Performance in Udacity Simulator}

In order to better evaluate and understand the relationship between both model accuracy and latency with respect to control performance, we conduct a systematic simulation study using 
the Udacity self-driving car simulator~\cite{udacitysim}. 

\begin{figure}[ht]
  \centering
  \includegraphics[width=.45\textwidth]{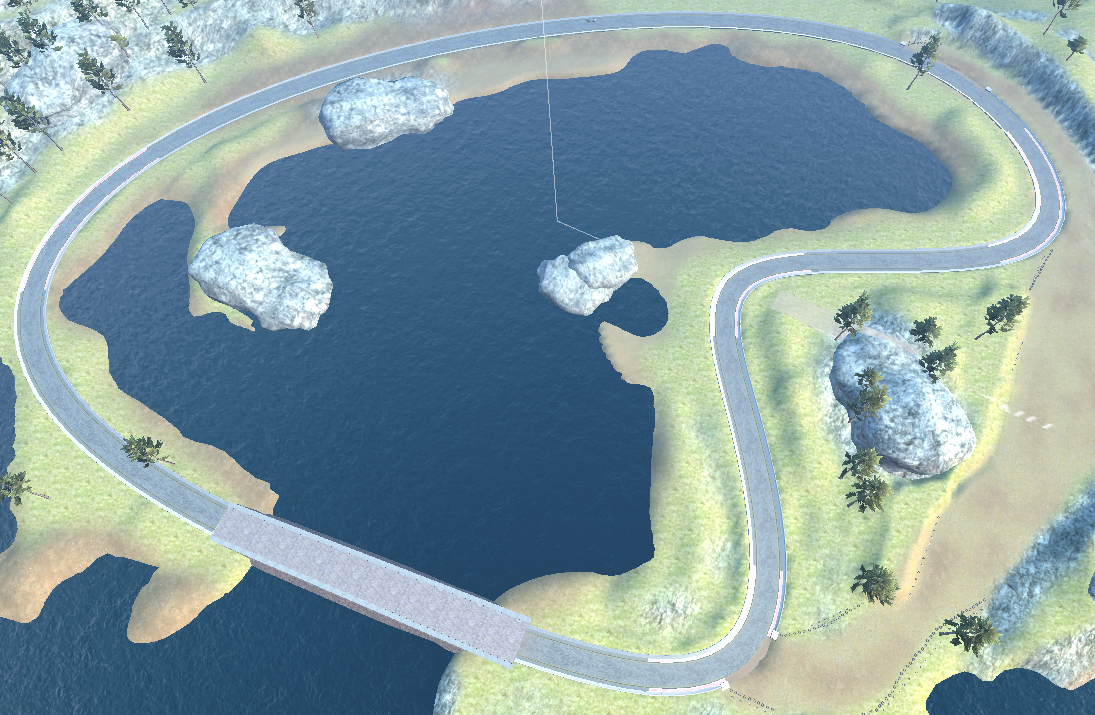}
  \caption{Udacity simulator's first default track.}
  \label{fig:track-sim}
\end{figure}

We run the simulator on a desktop computer that is equipped with a Nvidia GTX 1060 GPU and is running Ubuntu 20.04 for its OS. 
Using the simulator, we evaluated various models that could meet the resource and latency constraints of the DeepPicarMicro (i.e., Raspberry Pi Pico MCU's constraints). 
For this, we perform the same general workflow to find and train CNN models as we did for the DeepPicarMicro. We first manually collect training data by driving the simulated car around the first default track available in the simulator. Figure~\ref{fig:track-sim} shows an overview of the track we use for our simulation environments.

In this case, we collect a dataset of 14,468 samples. We then perform the same NAS approach as in Section~\ref{sec:nas}, 
and obtain models with varying validation losses.
In total, we train 84 models with validation losses ranging from 0.0259 to 0.0651.

We next evaluate the impacts that both validation loss and inference latency have on control performance. To accomplish this, we select a subset of six models with varying validation losses. For each model, we then add synthetic delays in the range of [0, 20, ..., 100] ms after each model inference to simulate longer model latencies and, by proxy, control actuations. Note that the actual inference times of the models on the PC were relatively negligible ($\sim$100-200$\mu$s). 
For each validation loss and inference latency combination, we adopt a similar methodology as in real-world experiments. This time, however, we measure how many seconds the car can drive before it crashes, more commonly known as the car's \textit{Time to Crash (TtC)}. Due to the increased size of the simulated track, we measure average TtC to better analyze control performance. For each test case, we measure the car's TtC value across five different runs, with each run having a maximum length of five minutes (300 seconds). Finally, we calculate the average TtC for each test case. This gives us a total of 30 data points.

\begin{table}[ht]
    \centering
    \begin{tabular}{|c|c|c|c|c|c|c|}
    \hline
         & &\multicolumn{5}{c|}{Validation Loss}  \\ \hline 
         \multirow{7}{*}{\rotatebox[]{90}{Latency (ms)}} 
         &      & 0.026    & 0.030    & 0.035    & 0.040    & 0.045    \\ \cline{2-7}
         & 0    & \col{57bb8a}300    & \col{57bb8a}300    & \col{f5e399}92     & \col{fde599}81     & \col{e67e74}29     \\ \cline{2-7}
         & 20   & \col{57bb8a}300   & \col{57bb8a}300    & \col{fbe488}84     & \col{ffe599}79     & \col{e78275}30     \\ \cline{2-7}
         & 40   & \col{57bb8a}300   & \col{57bb8a}300    & \col{fde599}81     & \col{fee498}78     & \col{e67c73}27     \\ \cline{2-7}
         & 60   & \col{57bb8a}300    & \col{57bb8a}300    & \col{ffe599}79     & \col{fddf97}76     & \col{e67c73}28     \\ \cline{2-7}
         & 80   & \col{f4e398}93     & \col{f5bc8a}59    & \col{f3b587}55     & \col{f6bf8b}60     & \col{e68074}29     \\ \cline{2-7}
         & 100  & \col{ed9c7e}43     & \col{f3b688}56     & \col{eb947b}39     & \col{eea180}45     & \col{e67d73}28     \\ \hline
    \end{tabular}
    \caption{Average TtC (seconds) performance for each test case on the Udacity simulator.}
    \label{tbl:sim-ttc}
\end{table}

\begin{table}[ht]
    \centering
    \begin{tabular}{|c|c|c|c|c|c|c|}
    \hline
         & &\multicolumn{5}{c|}{Validation Loss}  \\ \hline 
         \multirow{7}{*}{\rotatebox[]{90}{Latency (ms)}} 
         &      & 0.026    & 0.030    & 0.035    & 0.040    & 0.045    \\ \cline{2-7}
         & 0    & \col{57bb8a}0.00    & \col{6fc18c}0.12    & \col{91c98f}0.28     & \col{b5d292}0.46     & \col{dadb95}0.64     \\ \cline{2-7}
         & 20   & \col{7fc58d}0.20    & \col{98cb8f}0.32    & \col{bad392}0.48     & \col{dedc96}0.66     & \col{ffe399}0.84     \\ \cline{2-7}
         & 40   & \col{a8cf91}0.40    & \col{c1d593}0.52    & \col{e3de96}0.68     & \col{fee098}0.86     & \col{f9c98f}1.04     \\ \cline{2-7}
         & 60   & \col{d1d994}0.60    & \col{eadf97}0.72    & \col{fedd97}0.88     & \col{f8c78e}1.06     & \col{f3b086}1.24     \\ \cline{2-7}
         & 80   & \col{fae398}0.80    & \col{fdd995}0.92    & \col{f7c48d}1.08     & \col{f2ad85}1.26     & \col{ed967d}1.44     \\ \cline{2-7}
         & 100  & \col{facf91}1.00    & \col{f6bf8c}1.12    & \col{f1aa84}1.28     & \col{ec937c}1.46     & \col{e67c73}1.64     \\ \hline
    \end{tabular}
    \caption{Heuristic scores (between 0 to 2) for each test case on the Udacity simulator.}
    \label{tbl:sim-heuristic}
\end{table}

Table~\ref{tbl:sim-ttc} shows the results from the simulator tests. As expected, we find that both validation loss and inference latency play a vital role in the simulated car's control performance. 
For instance, the model with the lowest validation loss (highest accuracy) fails to stay on track if the inference latency is over 80 ms. As validation loss increases, this reduction in control performance (lower TtC) becomes even more apparent. Likewise, models with validation losses $\ge$0.035 also fail to remain on track, and crash before the allotted five minutes.

To validate our joint optimization strategy, we calculate the heuristic scores for each of the simulator test cases. 
In this case, 
we normalize the validation losses for all 84 models found in our NAS of the simulator dataset, as well as the synthetic delays that we added in our testing. 

Table~\ref{tbl:sim-heuristic} shows the heuristic scores for the 30 test cases. Similar to the real-world experiments, we find that the function does a relatively good job of predicting relative control performance. That being said, the function does not directly correlate to the control performance for the test cases, meaning that there is indeed room for improvement in our strategy. We leave this for future work. 
\section{Related Work}\label{sec:related}

There are several RC-car based autonomous car testbeds. 
MIT's RaceCar~\cite{shin2017project} and the F1Tenth car~\cite{upennf1tenth} are both based on a Traxxas 1/10 scale RC car.
Similar to the DeepPicar, the DonkeyCar also employs an end-to-end CNN-based control loop that runs on an embedded Raspberry Pi platform~\cite{donkeycar}.
The development of small-scale autonomous vehicle testbeds has also been seen in industry. 
For example, Amazon developed its own autonomous 1/18th scale RC car platform called DeepRacer~\cite{deepracer}.
However, all of these platforms employ microprocessor class computing platforms for their computational needs. 
In this paper, we instead introduce and evaluate of an MCU-based autonomous vehicle testbed.

In terms of the PilotNet architecture we use in this paper~\cite{Bojarski2016}, 
there has been work to improve its performance for autonomous vehicles~\cite{pilotnet20}. 
This includes a new architecture that utilizes a combination of residual layers, convolutional layers, and fully connected layers. Recent experiments also explored many avenues to improve performance, including data collection, pre-processing, and the use of a real-world representative simulator.
In our case, we use the original PilotNet architecture due to its popularity and simplicity, but plan to evaluate the newer iterations of PilotNet in future work.

With the goal of executing complex DNN-based algorithms on MCUs and other Edge devices, there has been a plethora of work in the TinyML sector~\cite{cai2018proxylessnas,cai2020once,cai2020tinytl,cai2021network,cai2022enable,lin2020mcunet,lin2021mcunetv2,banbury2021micronets,banbury2020benchmarking,wang2020apq}. Due to the relative infancy of the field, though, some works have focused on developing standards that can be used for benchmarking future works. For example, the TinyMLPerf benchmark suite was introduced in order to better enable TinyML-focused research~\cite{banbury2020benchmarking}. In addition, many machine learning frameworks have been developed that target MCUs. Apart from TFLM~\cite{david2021tensorflow}, there are other frameworks like CMSIS-NN~\cite{lai2018cmsis}, and uTensor~\cite{utensor}. In academia, the MCUNet framework has found much success in optimizing neural network discovery and inferencing on MCUs~\cite{lin2020mcunet}, achieving SOTA performance on image classification tasks by proposing an intelligent NAS approach and a highly optimized custom ML runtime. They have since extended this work to the MCUNetV2 framework, which instead prioritizes and optimizes peak memory usage for CNN models, thus allowing even bigger models to be deployed on MCUs~\cite{lin2021mcunetv2}. 
In our work, we adopted best-practices in TinyML research and applied them to CNN based end-to-end control of MCU-based autonomous CPS..
\section{Conclusion}\label{sec:conclusion}

We presented DeepPicarMicro, an autonomous RC car platform, which employs a deep-learning based end-to-end control on a tiny MCU.
We applied several DNN optimization techniques to execute the well-known PilotNet CNN architecture, which was used to drive NVIDIA's real self-driving car, on the platform's MCU. 
We also applied a state-of-the-art network architecture search (NAS) approach to find further optimized networks that can effectively control the car in real-time on the MCU. 
From an extensive systematic experimental and simulation study, we observed an interesting relationship between the accuracy, latency, and control performance of a system. Based on the insights, we proposed a joint optimization strategy that takes both accuracy and latency of a model in the network architecture search process for AI enabled CPS. 

For future work, we plan to evaluate more complex state-of-the-art CNN architectures on various MCUs. 
We also plan to investigate more fine-grained methods for estimating real-world control performance of AI enabled CPS systems and develop effective optimization strategies for MCUs.
\section*{Acknowledgements} \label{acknowledge}

This research is supported in part by NSF grant CNS1815959, CPS-2038923 and NSA Science of Security initiative contract no. H98230-18-D-0009.

\bibliographystyle{abbrv} 
\bibliography{reference}

\end{document}